\documentclass[journal]{IEEEtran}

\ifCLASSINFOpdf
\else
\fi

\usepackage{epsfig} 
\usepackage[nospace]{cite} 
\usepackage{amsmath,amssymb} 
\usepackage{algorithm,algpseudocode,algorithmicx}
\usepackage[mathscr]{eucal}
\usepackage{amsfonts,amssymb}
\usepackage{graphicx,subfigure,cases,multirow,graphics}
\usepackage[colorlinks,citecolor=blue,urlcolor=black,linkcolor=blue]{hyperref}
\usepackage[mathlines,switch]{lineno}
\usepackage{threeparttable}
\usepackage{booktabs}%

\newlength\savewidth

\newlength\savedwidth

\DeclareMathOperator\Per{Per}

\DeclareMathOperator\diver{div}
\DeclareMathOperator\Lip{Lip}
\DeclareMathOperator\MCRG{\textit{MCRander}}
\DeclareMathOperator\ADC{\textit{AdapCut}}
\DeclareMathOperator\VotMP{\textit{VotMaskProp}}

\newcommand{\fb}{\mathbf b}
\newcommand{\fc}{\mathbf c}
\newcommand{\fx}{\mathbf x}
\newcommand{\fy}{\mathbf y}
\newcommand{\fs}{\mathbf s}
\newcommand{\fw}{\mathbf w}
\newcommand{\fv}{\mathbf v}
\newcommand{\cC}{\mathcal C}
\newcommand{\cI}{\mathcal I}
\newcommand{\cG}{\mathcal G}
\newcommand{\cS}{\mathcal S}
\newcommand{\cR}{\mathcal R}
\newcommand{\cL}{\mathcal L}
\newcommand{\cM}{\mathcal M}
\newcommand{\cE}{\mathcal E}
\newcommand{\cV}{\mathcal V}
\newcommand{\cT}{\mathcal T}
\newcommand{\cU}{\mathcal U}
\newcommand{\cP}{\mathcal P}
\newcommand{\bR}{\mathbb R}
\newcommand{\sT}{\mathscr T}
\newcommand{\sV}{\mathscr V}
\newcommand{\kS}{\mathfrak S}

\newtheorem{remark}{Remark}

\begin{document}
%
\title{Mask Proposal Voting Based on  Geodesic Framework for Robust Image Segmentation}


\author{\IEEEauthorblockN{Li~Liu,  Mingzhu~Wang, Zhenjiang~Li, Da~Chen and Laurent~D.~Cohen,~\IEEEmembership{Fellow,~IEEE}}
\IEEEcompsocitemizethanks{
\IEEEcompsocthanksitem Li~Liu is with Yuanshen Rehabilitation Institute, Shanghai Jiao Tong University School of Medicine, 200025 Shanghai, China. (e-mail: liuli184791@sjtu.edu.cn) 
\IEEEcompsocthanksitem Mingzhu~Wang is with Yueyang Hospital of Integrated Traditional Chinese and Western Medicine, Shanghai University of Traditional Chinese Medicine, 200437 Shanghai, China. (e-mail:wangmingzhu@shutcm.edu.cn)
\IEEEcompsocthanksitem Zhenjiang~Li is with Department of Radiation Oncology, Shandong Cancer Hospital and Institute, Shandong First Medical University, Shandong Academy of  Medical Sciences, 250117 Jinan, China. (e-mail:zhenjli1987@163.com)
\IEEEcompsocthanksitem Da~Chen and Laurent~D.~Cohen are with University Paris Dauphine, PSL Research University, CNRS, UMR 7534, CEREMADE, 75775 Paris, France. (e-mail:chenda@ceremade.dauphine.fr, cohen@ceremade.dauphine.fr)}
}

\maketitle

%



\IEEEtitleabstractindextext{%
\begin{abstract}
Despite great advances, finding accurate segmentation remains a challenging task, especially in scenarios with cluttered backgrounds, complex intensity variations and topology appearance. Minimal path models have exhibited their strong ability in addressing image segmentation tasks. However, the performance of minimal paths-based segmentation approaches is heavily influenced by model initialization, hence limiting their application scope in practice. In this work, we propose a novel mask proposal voting framework that overcomes the major drawback of classical approaches,  allowing robust segmentation even in complicated scenarios.
Firstly, we introduce an efficient method for constructing  adaptive domain cuts as a constraint for initializing the  region-based min-cut evolution, by which diverse and reliable mask proposal candidates can be generated, substantially increasing the possibility of accurately covering the objective region by these proposals. Secondly, we propose a new mask voting scheme to build a voting score map encoding the final segmentation information. In contrast to classical path voting methods,  our model allows incorporating priors to assign different importance to each individual mask. As a consequence, the proposed segmentation model is  capable of accurately delineating object boundaries under complex scenarios, and is insensitive to initialization. Experiments  demonstrate that our method consistently outperforms state-of-the-art minimal path-based approaches in both accuracy and robustness.
\end{abstract}

}

\IEEEdisplaynontitleabstractindextext

%
\IEEEpeerreviewmaketitle

\section{Introduction}

\IEEEPARstart{I}{mage} segmentation is a fundamental task in artificial intelligence, computer vision and medical image analysis. In the past years, a great variety of approaches have been investigated for addressing the various segmentation problems occurred in complex scenarios~\cite{minaee2022image}. Significant examples involve segmentation approaches that integrate geometric models and partial differential equation (PDE) framework,  the data-driven segmentation models founded by the deep learning technique~\cite{falk2019u,ma2024segment,mazurowski2023segment}, as well as the combination of both ways~\cite{minaee2022image,ngan2018reformulating,marcos2018learning,ali2020end}.

It is known that variational approaches are usually established in the PDE framework and  have proven their strong ability in finding suitable solutions in a variety of tasks~\cite{kass1988snakes,peyre2010geodesic}. Among them, minimal path models, also regarded as  geodesic models, have obtained successful applications in many segmentation problems, due to its global optimization of the path energy, flexible framework that can involve types of image features and geometric features, as well as stable and fast numerical methods~\cite{sethian1999fast,mirebeau2014anisotropic,mirebeau2014efficient,mirebeau2019hamiltonian}. 
Computing optimal paths between two points can be back-tracked to Dijkstra's shortest path model~\cite{dijkstra1959note} under a discrete setting. However, this model usually suffers from a metrication error and cannot take into account geometric priors. Those shortcomings very often lead to unexpected segmentation results, especially in the extraction of curvilinear structures. Cohen and Kimmel proposed  a minimal path model~\cite{cohen1997global}, which treats an optimal continuous curve as the global minimizer of a geometric path energy in a PDE domain. The minimal path models are able to find the global minimum of the path energy, and have well-established numerical solvers~such as the fast marching methods~\cite{sethian1996fast,mirebeau2014anisotropic,mirebeau2014efficient,mirebeau2019riemannian}, thus have been widely applied to solve various image segmentation tasks.


\begin{figure*}[t]
\centering
\includegraphics[width=0.95\textwidth]{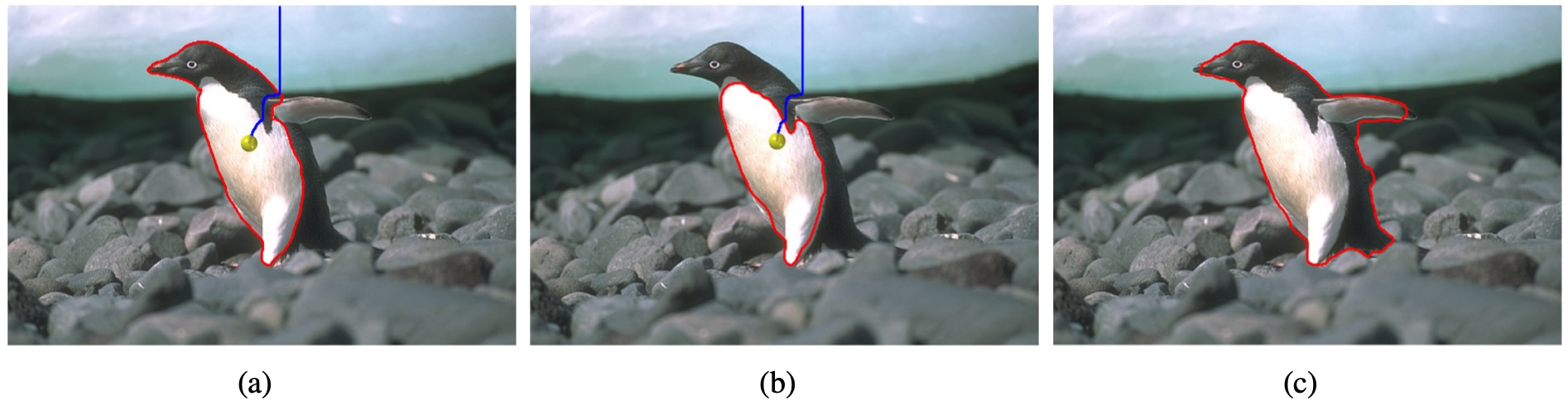}
\caption{Illustration of the comparison of different image segmentation methods on the image from the dataset~\cite{martin2001database}. The segmentation results (red lines) from the asymmetric geodesic voting model~\cite{zhou2025generalized}, the  min-cut Randers model~\cite{chen2024randers} and the proposed MPV model are exhibited in figures~\textbf{a} to \textbf{c}, respectively. The yellow dots indicate the landmark points inside the target region and the blue lines depict the adaptive domain cut~\cite{liu2024grouping}.}
\label{fig_Example}	
\end{figure*}

In the context of image segmentation, minimal paths are usually considered as an efficient tool for modeling the boundary of target regions. A great variety of interesting models~\cite{mille2015combination,appia2011active,kaul2012detecting} have been devoted to the construction of simple closed contours using a set of disjoint minimal paths, since in essence a minimal path is an open curve between the prescribed endpoints. Among these approaches, the core lies at the detection of saddle points from  geodesic distance map. An alternative way, named the circular geodesic model~\cite{appleton2005globally}, is to compute closed minimal paths whose endpoints are set as the identical one. This is done by using one source point located at the target boundary and a cut that connects a  point inside the target boundary to a point at the boundary of the image domain, and simultaneously passing through the source point. In the original circular geodesic model~\cite{appleton2005globally}, the cut is fixed as a straight ray-line, which may yield unexpected segmentation results when the cut intersects with the target boundary twice.  Chen~\emph{et al}.~\cite{chen2021geodesic} proposed a dual-cut geodesic computation scheme that utilizes a landmark point within the target region, where a simple closed contour is formed by combining two geodesic paths. Liu~\emph{et al}.~\cite{liu2024grouping} introduced an adaptive cut-based circular optimal path computation strategy, in which the closed contour consists of selected boundary proposals connected by geodesic paths. The adaptive domain cut (ADC) from the landmark point imposes a disconnection constraint on the image domain. Most of the minimal paths-based segmentation approaches as mentioned above exploit the image gradients as the features to define the target boundary. However, this very often yields unsatisfactory segmentation results since image gradients are local features and easy to be blurred by image noise. In contrast, a region-based Randers geodesic model~\cite{chen2024randers} was introduced to  overcome  the shortcoming induced by the locality of the image gradients-based features. By establishing the connection between the Randers HJB PDE and the regional appearance models, the nonlocal region-based features can be taken into account the computation of minimal paths.

The common idea in the classical geodesic voting methods (e.g.~\cite{rouchdy2013geodesic,zhou2025generalized})  is to   build a voting score map using a collection of piecewise geodesic paths, where the voting scores characterize the object boundary features, from which the segmentation contours can be recovered. In other words,  the voting score value at each point is derived by counting how many times of minimal paths passing through this point. The voting process allows reducing the probability of shortcuts or leakage issues in some extent. 
However, the classical geodesic voting methods typically utilize image gradients-inspired piecewise geodesic paths as voting elements for capturing boundary evidence. This restricts the voting process to sparse boundary locations, thus neglecting region-based homogeneity appearance features and prevents the model from leveraging interior region-based cues. Moreover, the absence of region-based features makes the geodesic voting scores which are highly sensitive to noise, complex image content and weakly-defined boundary features. 

In this paper, we introduce a novel mask proposal voting (MPV) model for robust image segmentation that integrates image region-based dissimilarity or similarity measurements and treats mask proposals as voting elements. A key ingredient of the MPV model is the computation of constrained adaptive domain cuts (ADCs), each constrained to pass through a prescribed boundary point. Mask proposals  are then generated using the  min-cut Randers geodesic model in conjunction with constrained ADCs. To construct reliable voting scores, we further introduce a weighted mask proposal voting scheme, in which mask proposals contribute under an unequal weighting strategy. 
We evaluate the proposed MPV model against state-of-the-art geodesic paths-based segmentation approaches on a real image, as illustrated in Fig.~\ref{fig_Example}. In this figure, the yellow point marks an interior point, the blue line denotes the adaptive cut~\cite{liu2024grouping}, and the red line indicates the resulting segmentation contour. Fig.~\ref{fig_Example}a shows the result of the asymmetric geodesic voting model~\cite{zhou2025generalized}, where part of the target region is missed due to intensity inhomogeneity. Fig.~\ref{fig_Example}b depicts the segmentation obtained with the min-cut Randers model~\cite{chen2024randers}, initialized from a circular geodesic path constructed using the domain cut and an isotropic minimal path. This method fails to accurately recover the target boundary, as it depends on initialization points close to the true boundary. In contrast, Fig.~\ref{fig_Example}c demonstrates that the proposed MPV model successfully delineates the complete target boundary.

%

The remainder of this paper is organized as follows. In Section~\ref{sec_Background}, we revisit the formulation of min-cut Randers geodesic model and of  the geodesic voting  framework. Section~\ref{Sec_Proposed} introduces the core contribution of this work: the mask proposal voting model, where the implementation details are presented in Section~\ref{Sec_Implement}. Experimental results and conclusions are provided in Sections~\ref{sec_Experi} and~\ref{Sec_Conclusion}, respectively.

\section{Background}
\label{sec_Background}


In this section, we first revisit the min-cut Randers model, which is regarded as an efficient tool for finding optimal contours and image segmentation. Then the classical geodesic voting method~\cite{rouchdy2013geodesic} that accumulates path evidence from a common source point are briefly introduced. 

\subsection{Revisiting of the Min-cut Randers Geodesic Model}
\label{subsec_Mincut}
\subsubsection{Min-cut Randers Model}
\label{subsec_RegionBasedMincut}
The region-based min-cut Randers geodesic model~\cite{chen2024randers} is developed to minimize the classical min-cut problem, or region-based active contours problem,  under the static HJB PDE frame. Typically, let $\Omega$ be an open bounded domain, $\cR\subset\Omega$  the foreground and $\Omega\backslash\cR$ the background. The foreground $\cR$ is also referred to as a \emph{shape}, whose boundary is represented as $\partial\cR$. The region-based min-cut problem aims to minimize the following functional
\begin{equation}
\label{eq_euclideanCase}
E(\cR)=\lambda\int_\cR \xi(\fx)d\fx+\Per(\partial \cR)
\end{equation}
where $\xi:\Omega\to\bR$ is a function of Lipschitz continuity,  $\Per(\partial\cR)$ is the perimeter of $\cR$ and $\lambda\in\bR^+$ is a weighting parameter. Let $\gamma_\cR:[0,1]\to\partial\cR$ be a regular curve that  parameterizes $\cR$ in a clockwise order, yielding a  reformulation of the regularization as the Euclidean length of  $\gamma_\cR$, i.e. $\Per(\partial \cR)= \int_0^1\|\gamma^\prime_\cR(t)\|dt$, where $\gamma^\prime_\cR$ is the first-order derivative of $\gamma_\cR$.
In~\cite{zach2009globally,chen2024randers}, the regularization term in Eq.~\eqref{eq_euclideanCase} is generated as a weighted curve length
\begin{equation*}
\cL_{\cM}(\gamma_{\cR})=\int_0^1\sqrt{\langle\gamma^\prime_\cR(t), \cM(\gamma_\cR(t))\gamma^\prime_\cR(t)\rangle}\;dt
\end{equation*}
where  $\cM(\fx)$ is a positive definite symmetric matrix of size $2\times2$. The matrices $\cM$ is constructed by the image gradients, involving both their directions and magnitude values~\cite{chen2024randers}.
The energy functional~\eqref{eq_euclideanCase} can be reformulated as
\begin{equation}
\label{eq_RiemannCase}
\cE(\cR)=\lambda\int_\cR\xi(\fx)d\fx + \cL_{\cM}(\gamma_{\cR}).
\end{equation}
Note that $\xi$ can be derived as the shape gradient of a region-based image appearance model, for instance, the piecewise constant models~\cite{chan2001active}, the histograms-based model~\cite{lleg2007Image} and the pairwise similarity/dissimilarity model~\cite{bertelli2008variational,jung2012nonlocal}. These models characterize the feature homogeneity measured respectively within the shape $\cR$ and the background $\Omega\backslash\cR$. The minimization of $\cE$ is usually implemented by iterating  the following steps: (i) fixing the shape $\cR$ to minimize the functional $\cE(\cR)$, yielding an optimal shape $\cR^*$, and (ii) updating the shape gradient $\xi$ by $\cR^*$.

\begin{algorithm}[t]
\caption{Min-cut Randers Model with A Cut} 
\label{alg_CirCurveEvol} 
\renewcommand{\algorithmicrequire}{\textbf{Input:}}
\renewcommand{\algorithmicensure}{\textbf{Output:}}
\begin{algorithmic}[1]
\Require A cut $\cC$, an initial curve $\gamma_0$ and a source point $\fs\in\cC$.
\Ensure The circular min-cut geodesic path $\gamma_\infty$ . 
\renewcommand{\algorithmicensure}{\textbf{Initialization:}}
\Ensure Set the index $j=0$.
\While{stopping criterion is not satisfied }
\State Update the tubular neighbourhood $\sT_j$ of $\gamma_j$;
\State Compute the vector field $\fw_j$  in $\sT_j$ using Eq.~\eqref{eq_DivSolver}.
\State Set $\omega_j(\fx)=\fw_j(\fx)^\perp$.
\State Minimizing the circular geodesic problem~\eqref{eq_CircularMinCut};
\State Track the circular min-cut geodesic path $\gamma_{j+1}$; 
\State Set $\gamma_\infty\leftarrow\gamma_{j+1}$;
\EndWhile 
\end{algorithmic}	
\end{algorithm}

\subsubsection{Evolving Randers Geodesic Paths}
\label{subsec_EvolvingMincut}
The min-cut Randers geodesic model~\cite{chen2024randers} transfers the energy~\eqref{eq_RiemannCase} to a weighted curve length  using the divergence theorem. Given a shape $\cR_j\subset\Omega$ whose boundary is parameterized by $\gamma_j:[0,1]\to\Omega$, one can estimate the corresponding shape gradient $\xi_j$ via $\cR_j$. In the spirit of iterative minimization scheme,  the functional~\eqref{eq_RiemannCase} in the $j$-th iteration for $j\geq 0$ is approximated as
\begin{equation}
\label{eq_RandersCase}
\cE_{j}(\gamma_\cR):=\lambda\int_0^1\left\langle\omega_j\bigl(\gamma_\cR(u)\bigr), 
\gamma_\cR^\prime(u)\right\rangle du+ \cL_{\cM}(\gamma_\cR)
\end{equation}
where $\omega_j$ is a vector field defined over a tubular region $\sT_j$ surrounding the input curve $\gamma_j$, i.e. $\omega_j:\sT_j\to\bR^2$. More specifically, one has $\omega_j(\fx)=\fw_j(\fx)^\perp$ where $\fw_j$ is computed by solving a PDE-constrained optimization problem~\cite{chen2024randers}:
\begin{equation}
\label{eq_DivSolver}
\min\int_{\sT_j}\|\fw_j(\fx)\|d\fx,\quad s.t.\quad \diver(\fw_j)=\xi_j\quad\text{on}\quad \sT_j,
\end{equation}
where $\diver$ is the divergence operator.

\subsubsection{Circular Geodesic Paths with a Straight Segment  Cut}
A geodesic path in essence is an open curve, whereas in the context of image segmentation one expects a simple closed curve to depict the target boundary. In~\cite{chen2024randers}, the authors regard an evolving contour as a circular path, whose computation relies on a domain cut $\cC\subset\Omega$. In particular, the domain cut $\cC$ is a \emph{straight segment} with two endpoints respectively located at the two boundaries of the tubular neighbourhood.

The domain cut $\cC$ serves as a wall with the tubular neighbourhood, such that its two sides are disconnected. This means that a path cannot pass through $\cC$. Hence the minimization of~\eqref{eq_RandersCase} is transferred as 
\begin{equation}
\label{eq_CircularMinCut}
\min_{\gamma_\cR\in\Lip([0,1],\Omega)}~\cE_{j}(\gamma_\cR),~s.t.~
\begin{cases}
\gamma_\cR(0)=\gamma_\cR(1)=\fs\\
\gamma_\cR(u)\notin\cC,\forall u\in(0,1)
\end{cases}	
\end{equation}
where $\fs$ is a source point that is located in the domain cut $\cC$. During the evolution, we assume that the source point $\fs$ is fixed. The evolution will be terminated once two successive contours are sufficiently close to each other~\cite{chen2024randers}. 
For the sake of simplicity, we define a function $\MCRG$ to transfer an initial curve $\gamma_0$ to an optimal curve using the min-cut Randers model with a domain cut, as described in Algorithm~\ref{alg_CirCurveEvol}. Specifically, the function $\MCRG$ is formulated as
\begin{equation}
\label{eq_MCRM}
\gamma_\infty=\MCRG(\gamma_0,\cC,f),
\end{equation}
where $\cC$ is the domain cut (straight segment) and  $f:\Omega\to\bR$ is a gray level image. During the evolution the shape gradient $\xi_j$ is derived from $f$ with a given image appearance model.

The min-cut Randers geodesic model formulates image segmentation as an optimal path computation problem that integrates region-based appearance cues with geometric regularization. Using a straight segment cut as an obstacle is computationally efficient for deriving circular minimal paths. However, this strategy is limited in localizing target regions, as it relies solely on the source point while ignoring interior and exterior location priors.

\begin{figure*}[t]
\centering
\includegraphics[width=0.95\linewidth]{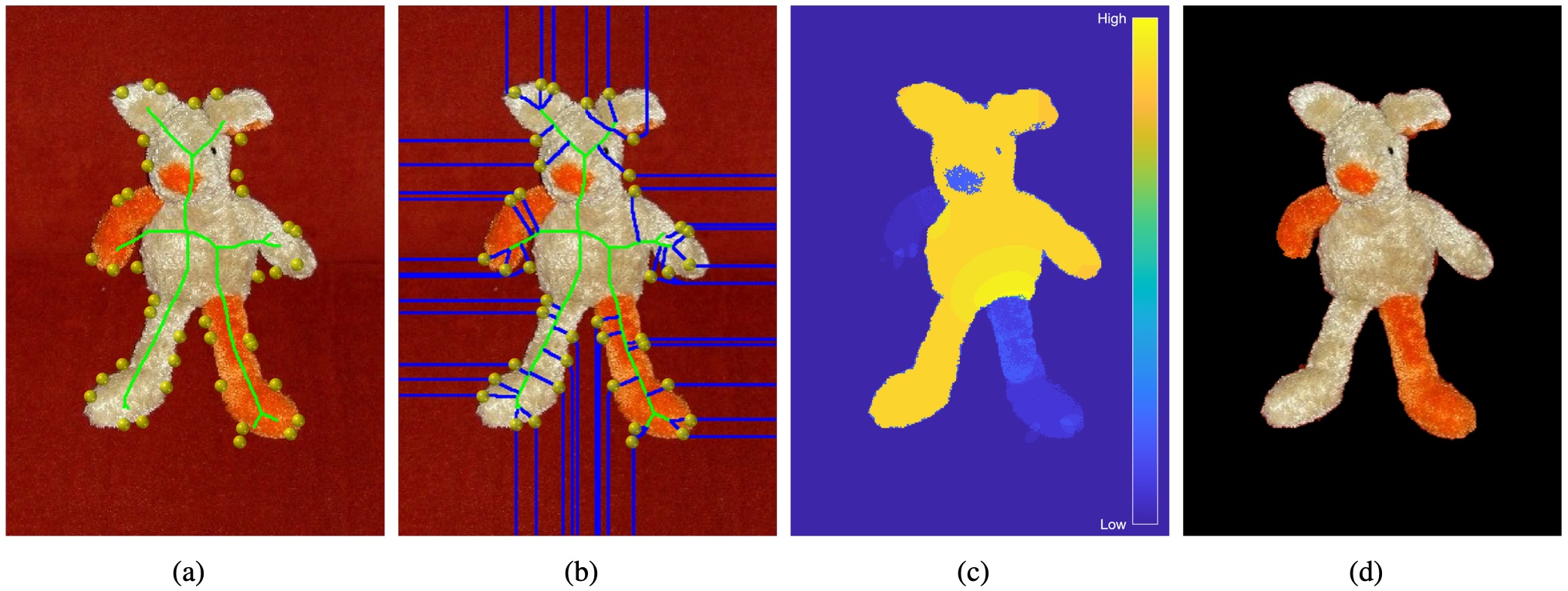} 
\caption{Overview of the proposed MPV model. (\textbf{a}) Illustration of the vertices and skeleton structure, which are  indicated  by yellow dots and green lines, respectively. (\textbf{b}) Visualization of the constrained ADCs. (\textbf{c}) illustration for the mask voting score map. (\textbf{d}) Segmentation result. The black color indicates the background. Note that in this figure we only demonstrate a part of the vertices $\fv_k$ and the image is from the dataset~\cite{rother2004grabcut}. }
\label{fig_OverviewFlow}	
\end{figure*} 

\subsection{Voting with Geodesic Path Elements}
\label{subset_GVS}

The objective of the classical geodesic voting method~\cite{rouchdy2013geodesic} is a voting score map $\cV_\fs:\Omega\to[0,\infty)$ associated to a source point $\fs$  to capture image characteristics. Basically, the map $\cV_\fs$ is generated from a set of geodesic paths $\Phi=\{\cG_j\}_{1\leq j\leq J}\subset\Lip([0,1],\Omega)$, where $J$ is a positive integer. This is to say
\begin{equation*}
\cV_\fs(\fx)=\sum_{\cG\in\Phi}\delta(\fx,\cG),
\end{equation*}
where $\delta:\Omega\times \Lip([0,1],\Omega)\to\{0,1\}$ is a detector defined being such that
\begin{equation}
\label{eq_Pathdetct}
\delta(\fx,\gamma)=
\begin{cases}
1,&\exists u \in[0,1],~\gamma(u)=\fx,\\
0,&\text{otherwise}.
\end{cases}	
\end{equation}
In other words, $\delta(\fx,\gamma)$ takes the value 1 when the curve  passes through the point $\fx$, and $0$ otherwise. 
 
 The voting map $\cV_\fs$ critically depends on the selection of endpoints, which can be typically chosen as the image domain boundary points~\cite{rouchdy2013geodesic}, prescribed edge points~\cite{ghorpade2015automatic}, or as farthest points~\cite{peyre2006geodesic}. Moreover, existing models predominantly employ a metric driven by image gradient information~\cite{zhou2025generalized,ghorpade2015automatic}. Importantly, conventional geodesic voting frameworks consider only the spatial trajectories of geodesic paths originating from a common source point and rely solely on edge features. To overcome these limitations, our approach incorporates region-based cues derived from the Min-cut Randers Geodesic Model into the voting scheme, thereby improving segmentation robustness in complex scenarios. As illustrated in Fig.~\ref{fig_Example}, the proposed model outperforms competing methods.

\begin{algorithm}[t]
\caption{Summary algorithm of the MPV Model} 
\label{algo_Overview}
\begin{algorithmic}
\renewcommand{\algorithmicrequire}{\textbf{Input:}}
\renewcommand{\algorithmicensure}{\textbf{Output:}}
\Require The image $f$, the set $\sV=\{\fv_k\}_k$, the initial contours $\{\Gamma_k\}_k$, and the skeleton structure $\cI$.
\Ensure  Voting score map $\cS$.
\renewcommand{\algorithmicrequire}{\textbf{Initialization:}}
\Require  For each point $\fx\in\Omega$, set $\cS(\fx)=0$.
\end{algorithmic}
\begin{algorithmic}[1]
\For{each vertex $\fv_k \in \sV$}
\State Compute an ADC $\cC_k=\ADC(\cI,\fv_k,\Gamma_k)$.
\label{algLine_ADC}
\State Compute a contour $\gamma_{0,k}$ by the ADC $\cC_k$ for initializing the min-cut Randers model.
\label{algLine_IniCircularPath}
\State Compute $\cG_k=\MCRG(\gamma_{0,k},\cC_k,f)$ using the min-cut Randers model with the ADC $\cC_k$.
\label{algLine_MCRG}
\State Compute a shape gradient $\tilde\xi_k$ using the contour $\cG_k$.
\label{algLine_SG}
\State Update  voting scores $\cS$ by $\cS\gets\VotMP(\tilde\xi_k,\cG_k)$. 
\label{algLine_Vote}
\EndFor
\end{algorithmic}
\end{algorithm}

\section{Mask Proposal Voting Model}
\label{Sec_Proposed}

In this section, we present the core of this work, the mask proposal voting (MPV) model, which consists of two main components: (i) an efficient strategy for generating mask proposals using constrained ADCs, and (ii) the construction of a voting score map that assigns different importance to mask proposals.

\subsection{Overview of the Proposed MPV Model}

Prior to introducing the details of each component, we provide an overview of the proposed model, as illustrated in Fig.\ref{fig_OverviewFlow} and summarized in Algorithm\ref{algo_Overview}. The initialization of the MPV model consists of a set of vertices $\sV={\fv_k}_{1\leq k \leq K}$ and a skeleton structure $\cI\subset\Omega$ (a set of connected points) of the foreground region $\cR$, where $K$ denotes the number of vertices. Fig.\ref{fig_OverviewFlow}a shows the vertices $\fv_k$ together with the skeleton $\cI$. In the context of image segmentation, these vertices are expected to be distributed densely near the target boundary and, together with the skeleton structure, provide sufficient information to construct a set of constrained ADCs, which form the foundation of the proposed model (see Fig.\ref{fig_OverviewFlow}b). Each constrained ADC subsequently yields a mask proposal, denoted by $\cR_j\subset\Omega$ for $j\geq 1$, obtained through an ADC-based min-cut geodesic model. Finally, the set of mask proposals is integrated within a weighted region-based voting scheme to construct a voting score map, from which the final segmentation is derived.

\begin{figure*}[t]
\centering
\includegraphics[width=0.95\textwidth]{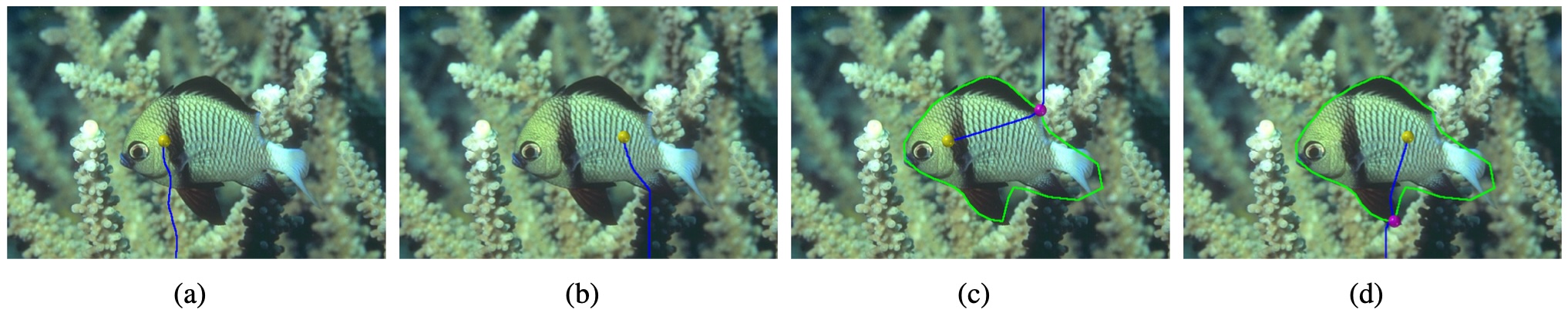}
\caption{Comparison of ADCs computed using~\cite{liu2024grouping} and the proposed method. (\textbf{a}) and (\textbf{b}) The blue lines represent ADCs which are generated from two different interior points (yellow dots) within the target region. (\textbf{c}) and (\textbf{d}) The blue lines illustrate the ADCs computed using the proposed method, each of which is constrained to pass through a specified vertex (magenta dot) on the prescribed contour (green line). The image is from the dataset~\cite{martin2001database}.}
\label{fig:cutExamp}	
\end{figure*}

In particular, Line~\ref{algLine_ADC} of Algorithm~\ref{algo_Overview} calls the function $\ADC$ to compute a constrained ADC, as detailed in Section~\ref{subsec_DomainCuts}. Line~\ref{algLine_IniCircularPath} computes a simple closed curve that serves as the initialization for the min-cut Randers model, (see Section~\ref{sub_RegProFusionVS}). In Line~\ref{algLine_MCRG}, the function 
$\MCRG$ computes optimal evolved paths using the min-cut Randers model, as described in Section~\ref{sub_RegProFusionVS}, and is used to generate mask proposals as voting elements in Line~\ref{algLine_SG}. In addition, Line~\ref{algLine_Vote} compute the voting scores, described in Section~\ref{subsec_Fusion}. Finally, the construction for the vertex set $\sV$ and the skeleton structure is described in Section~\ref{Sec_Implement}.

\subsection{Construction of a Constrained ADC}  
\label{subsec_DomainCuts}
In this section, we firstly revisit the computation of the classical ADC and then analyze its shortcomings,  to clearly  highlight the advantages of the introduced constrained ADCs.

\subsubsection{Classical ADC and its Shortcomings}
In~\cite{liu2024grouping}, an ADC is regarded as a geodesic path connecting an interior point of the target region to a point $\fx\in\partial\Omega$, i.e. an ADC is modeled as an optimal curve $\cC\in\Lip([0,1],\Omega)$ s.t. $\cC(0)\in\cI$ and $\cC(1)\in\partial\Omega$. 
The implementation can be carried out by the eikonal PDE framework~\cite{cohen1997global}, relying on a geodesic distance map:
\begin{equation*}
\mathfrak U_\cI(\fx)=\min_{\gamma\in\Lip([0,1],\Omega)}~\int_0^1 \psi(\gamma)\|\gamma^\prime\|dt,~s.t.~
\begin{cases}
\gamma(0)\in\cI\\
\gamma(1)=\fx,	
\end{cases}
\end{equation*}
where $\psi:\Omega\to\bR^+$ is a potential. It can be defined by 
\begin{equation}
\label{eq_CutPotential}
\psi(\fx)=\exp\big(\tau g(\fx)\big)+w,
\end{equation}
where $\tau,\,w\in\bR^+$ are positive constants and  $g(\fx)$ is the magnitude of the image gradient at $\fx$. The geodesic distance map $\mathfrak U_\cI$ is the unique viscosity solution to the eikonal PDE
\begin{equation}
\label{eq_IsoEikonalPDE}
\|\nabla\mathfrak{U}_\cI(\fx)\|=\psi(\fx),\quad\forall\fx\in\Omega\backslash\cI\\
\end{equation}
with boundary condition $\mathfrak{U}_\cI(\fx)=0,\forall \fx\in\cI$. 


In the classical circular geodesic models~\cite{appleton2005globally,liu2024grouping}, a cut usually provides the source point (which is also the endpoint) of the circular geodesic path, such that the circular paths are heavily determined by the cut. However, a major shortcoming of the classical ADC is the insufficient constraint on controlling its location and shape, since the computation of the classical ADC strongly relies on the potential $\psi$. As a result, the classical ADC lacks reliable geometric constraints for enhancing the segmentation.  To overcome this issue, we propose a new constrained ADC, such that a satisfactory ADC should pass by a vertex $\fv_k\in\Omega$. Fig.~\ref{fig:cutExamp} illustrates examples for the comparison between the classical ADC (Figs.~\ref{fig:cutExamp}a and \ref{fig:cutExamp}b) and the proposed constrained ADC (Figs.~\ref{fig:cutExamp}c and \ref{fig:cutExamp}d). 

\subsubsection{ Construction of Constrained ADCs}
\label{subsubSec_ADC}
The expected constrained ADCs are required to originate from the skeleton $\cI$,  pass through a prescribed vertex $\fv_k\in\sV$, and terminate at the boundary $\partial\Omega$. A straightforward approach would be to construct two minimal paths with  $\fv_k$ as their common source point: one connects $\fv_k$ to $\cI$, and another connects $\fv_k$ to $\partial\Omega$. However,  the concatenation of those two paths may cross the target boundary multiple times, leading to undesirable segmentation results. In contrast,  we propose an alternative strategy for computing satisfactory ADCs in conjunction with a prescribed segmentation contour.

The basic idea is to construct a circular obstacle between the image domain boundary $\partial\Omega$ and the skeleton $\cI$, so that the obstacle  passes through the vertex $\fv_k$. We then regard $\fv_k$ as a ``hole'' on this obstacle. From the perspective of the minimal path framework, the geodesic distance front will originate from the skeleton $\cI$, propagate through the vertex $\fv_k$, and terminate once any point at $\partial\Omega$ is reached. For this purpose, we compute a simple closed contour $\Gamma_k$ which passes through $\fv_k$,  encloses the skeleton $\cI$, and remains  disjoint from $\partial\Omega$. Such a contour $\Gamma_k$ can be easily generated, as described in Section~\ref{Sec_Implement}.
Our objective is to extract an open optimal path $\cC_k$ with $\cC_k(0)\in\cI\text{~and~}\cC_k(1)\in \partial\Omega$, and 
\begin{equation}
\begin{cases}
\exists u\in(0,1),~\cC_k(u)=\fv_k\\
\forall u\in[0,1],~\cC_k(u)\notin \Gamma_k\backslash\{\fv_k\}.
\end{cases}
\end{equation} 
Let $\cT(\Gamma_k)=\{\fx\in\Omega,\;\min_{\fy\in\Gamma_k}\|\fx-\fy\|<\tau \}$ denote the tubular neighborhood of $\Gamma_k$, where $\tau$ specifies the maximal width.  To mitigate the influence from irrelevant image content, we  restrict the potential definition to image data within $\cT(\Gamma_k)$: 
\begin{equation}
\label{eq_CutPotential2}
\psi_k(\fx)=
\begin{cases}
\psi(\fx),&\text{if~}\fx\in \cT(\Gamma_k) \\
\epsilon,&\text{otherwise}
\end{cases}
\end{equation}
 where $\epsilon>0$ is a constant. The tubular neighborhood thus enforces spatial locality by excluding image data far away from the prescribed contour, while simultaneously encouraging the tangent of the computed ADC near the specified vertex to align closely with the direction orthogonal to the boundary.

Similar to the classical ADC, the proposed constrained ADC is computed by incorporating the new potential $\psi_k$ into the eikonal PDE framework~\cite{cohen1997global}, with the constraint imposed by $\tilde\Gamma_k:=\Gamma_k\backslash \{\fv_k\}$. Specifically, the geodesic distance map $\cU_k$ is defined as
\begin{align}
\label{eq_ConstrainedDistance}
\cU_k(\fx)= &\min_{\gamma\in\Lip([0,1],\Omega)} 
      \int_0^1 \psi_k(\gamma(u)) \|\gamma^\prime(u)\|du,\nonumber\\  
 &\text{subject~to}~
\begin{cases}
 \gamma(0)\in\cI,~\gamma(1)=\fx, \\
 \gamma(u)\notin\tilde\Gamma_k,~\forall u\in[0,1].
\end{cases}
\end{align}

The distance map~\eqref{eq_ConstrainedDistance} can be numerically estimated using the classical Fast Marching method~\cite{sethian1996fast,mirebeau2019riemannian} through  single-pass wavefront propagation. In this process, the contour $\tilde\Gamma_k$ serves as an obstacle~\cite{chen2023geodesic}, halting the advance of the wavefront, while the vertex $\fv_k$ plays a role of  a ``hole", permitting the wavefront to pass through. Finally, an endpoint $\fb$ of the constrained ADC is the point at the image domain boundary $\partial\Omega$ with minimum distance $\cU_k$. The constrained ADC can be computed via a gradient descent procedure on the map $\cU_k$ from the detected point $\fb$ till a point in the set $\cI$ is reached~\cite{cohen1997global}. 

\begin{figure*}[t]
\centering
\includegraphics[width=0.95\textwidth]{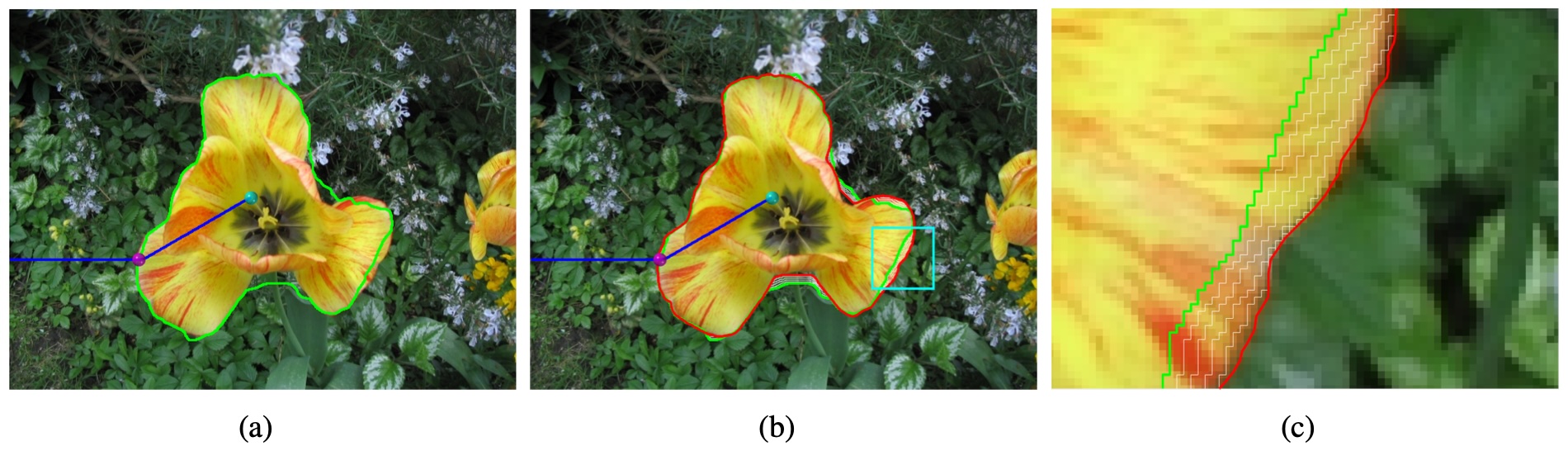}
\caption{Illustration of the contour evolution process for extracting the circular min-cut geodesic path. (\textbf{a}) Initialization using a circular geodesic path (green line) derived from the orientation-constrained domain cut (blue line). The cyan point marks the landmark position, and the magenta point indicates the specified vertex. (\textbf{b}) Intermediate contour evolution steps (white lines) from the initial contour (green line) to the final segmentation result (red line). (\textbf{c}) Enlarged view of the subregion highlighted by the cyan rectangle in Figure~\textbf{b}. The image is from the dataset~\cite{rother2004grabcut}.}
\label{fig:evolution}	
\end{figure*}

\subsection{Construct Mask Proposals via  Constrained ADCs}
\label{sub_RegProFusionVS}
As discussed in~\cite{chen2024randers}, combining the region-based appearance features with image gradients significantly improves foreground-background segmentation over image gradients-only models. Motivated by this, we 
apply the min-cut Randers model in conjunction with the introduced constrained ADCs. 

A simple closed curve is required to initialize  the min-cut Randers model, as described in Section~\ref{subsec_Mincut}. 
Recall that the input of the MPV model is a set of vertices $\sV$ and a skeleton structure $\cI$. 
Using the method introduced in Section~\ref{subsubSec_ADC}, each vertex $\fv_k\in\sV$ generates a constrained ADC $\cC_{k}$ connecting $\cI$ to $\partial\Omega$.  We then solve the following circular geodesic problem:
\begin{align}
\label{eq_circularMP}
&\min_{\gamma\in\Lip([0,1],\Omega)} 
   \int_0^1 \cP(\gamma(u)) \|\gamma^\prime(u)\|du, \\
&\text{subject to~}
\begin{cases}
  \gamma(0)=\gamma(1)=\fv_k, \\
  \gamma(u)\notin \cC_k\cup\cI,~\forall u\in(0,1),
\end{cases}\nonumber
\end{align}
where $\cP:\Omega\to\bR^+$ is a potential defined as a decreasing function of the image gradient magnitude. 
Solving~\eqref{eq_circularMP} yields a circular path $\tilde\cG_{k}$ that intersects with $\cC_k$ at the vertex $\fv_k$, thereby enclosing $\cI$.  The circular path $\tilde\cG_{k}$ and the ADC $\cC_k$ are then used, as the initial contour and obstacle respectively,  to initialize the  min-cut Randers geodesic model, as detailed in Section~\ref{subsec_Mincut} and Algorithm~\ref{alg_CirCurveEvol}. 
The contour evolution process for extracting the circular path is illustrated in Fig.~\ref{fig:evolution}.

We denote by $\cG_k$ the final contour obtained by the min-cut Randers model initialized with $\tilde\cG_k$. Its interior region, denoted by $\kS_k\subset\Omega$, is regarded as the mask proposal associated with  the vertex $\fv_k$. For the detailed numerical solution to problem~\eqref{eq_circularMP} and to the min-cut Randers model, we refer the reader to~\cite{chen2023geodesic,chen2024randers}.

\subsection{Voting with Mask Proposal Elements}
\label{subsec_Fusion}
In this section, we describe the construction of the voting score map $\cS:\Omega\to[0,\infty)$ using mask proposals as voting elements. The goal is to estimate, for each $\fx\in\Omega$, the likelihood of belonging to the target region, thereby characterizing the region of interest. Unlike classical geodesic voting strategy that relies on sparse boundary cues, our method leverages mask proposals to incorporate both boundary- and region-based features. By aggregating these cues, the proposed method produces a voting score map that is both denser and more robust.
Formally, the voting score map is defined as
\begin{equation}
\label{eq:VSmap}
\cS(\fx)=\sum_{1\leq k \leq K}\eta(\fx~|~\xi_k,\kS_k),
\end{equation}
where $\eta$ is a detector given by
\begin{equation}
\label{eq_Regiondetct}
\eta(\fx~|~\xi_k,\kS_k)=
\begin{cases}
\alpha_1,& \text{if}~\xi_k(\fx)\chi_{\kS_k}(\fx)<0,\\
\alpha_2,& \text{if}~\xi_k(\fx)\chi_{\kS_k}(\fx)>0,\\
0,&\text{otherwise}.
\end{cases}	
\end{equation}
with $\alpha_1>\alpha_2\geq 0$ denoting two constants, and $\chi_{\kS_k}$ the indicator function of the mask proposal $\kS_k$. As discussed in~\cite{chen2021generalized}, the sign of the shape gradient $\tilde\xi_k$ provides information about the target region: $\tilde\xi_k(\fx)>0$ indicates that $\fx$ lies outside the region, $\tilde\xi_k(\fx)=0$ corresponds to the boundary, and $\tilde\xi_k(\fx)<0$ denotes the interior. The formulation in~\eqref{eq_Regiondetct} integrates cues from region-based image features and the generated mask proposals, thereby facilitating accurate segmentation even in challenging scenarios. It is worth noting that the image appearance model used to compute the shape gradient $\tilde\xi_k$ may differ from that adopted in the min-cut Randers model. Intuitively, such an unequal weighting ensures that reliable mask proposals contribute more strongly to $\cS$, while noisy or inconsistent ones are naturally suppressed.

\begin{algorithm}[!t]
\caption{Generation of the Vertex Set $\sV$} 
\label{algo_Implementation}
\begin{algorithmic}
\renewcommand{\algorithmicrequire}{\textbf{Input:}}
\renewcommand{\algorithmicensure}{\textbf{Output:}}
\Require A raw image  defined in the domain $\Omega$.
\Ensure  A vertex set $\sV$.
\end{algorithmic}
\begin{algorithmic}[1]
\State Predict a vertex set $\{\fy_\ell\}$ and a mask $\kS_0$ by PolarMask.
\State Use the mask $\kS_0$ to generate the skeleton structure $\cI$.
\renewcommand{\algorithmicrequire}{\textbf{Initialization:}}
\Require Set $\sV=\{\fy_\ell\}_{1\leq \ell \leq L}$.
\For{each predicted vertex $\fy_\ell$}
    \State Compute an ADC by $\cC_\ell= \ADC(\cI,\fy_\ell,\partial\kS_0)$.
    \State Compute a circular path $\cG_\ell$ with $\cC_\ell$ by Eq.~\eqref{eq_circularMP}.
    \State Sample a set of points $\fv_{\ell,i}$ for $i>0$ from $\cG$.
    \State Update the set $\sV\gets\sV\cup\{\fv_{\ell,i}\}_i$. 
\EndFor
\end{algorithmic}
\end{algorithm}

\section{Implementations}
\label{Sec_Implement}

In this section, we describe the implementation for constructing the vertex set $\sV=\{\fv_k\}_{1\leq k \leq K}$ and the skeleton $\cI$. To leverage the strong representation capability of deep learning techniques~\cite{minaee2022image}, we employ the PolarMask model~\cite{xie2021polarmask++} to compute  $\cI$ and to  initialize the construction of $\sV$.  As a simple yet effective instance segmentation method, the PolarMask model incurs negligible computational overhead  while maintaining high efficiency, making it  a suitable pre-segmentation module in our work. It is worth mentioning that  other segmentation models~(e.g.~\cite{10599205,cao2022swin,10460426,10436544}) based on deep learning techniques can also be applied. 

The PolarMask model represents the object contour in polar coordinates. Specifically, the contour is parameterized by a center point $\fc$ and a set of ray-line segments which emanate from $\fc$ and extend outward to the object boundary. The endpoints of those ray-line segments are the vertices $\fy_\ell$ for $1\leq \ell \leq L$ with $L$ a positive integer.  The boundary of the predicted mask $\kS_0$ is then represented as a polygon by sequentially connecting  vertices  $\fy_\ell$. We apply a series of mathematical morphology operators to erode the predicted mask $\kS_0$ to shrink the object boundaries, encouraging the resulting mask to be contained within the target region. Subsequently, a skeletonization procedure is performed to generate the skeleton structure $\cI$ that preserves the topological properties of the object. This yields a compact set of structurally consistent landmark points, which serve as source points for the computation of the constrained ADCs.

\begin{figure*}[h]
\centering
\includegraphics[width=0.95\textwidth]{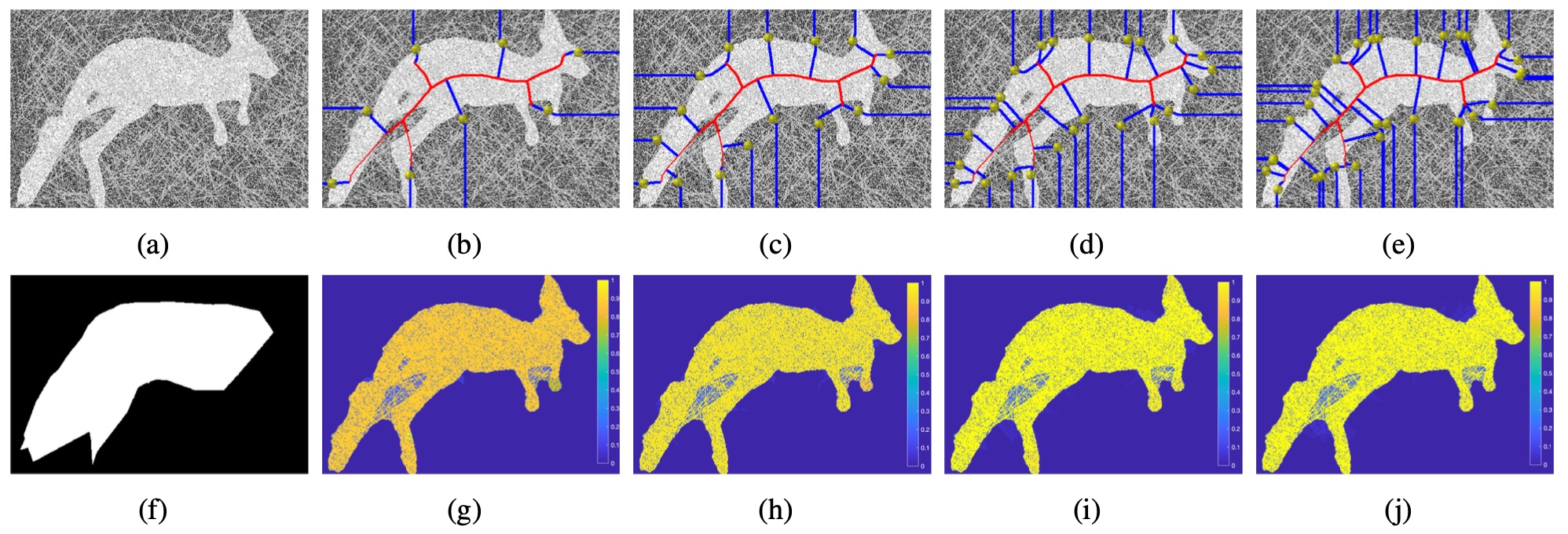}
\caption{Effect of varying the number of vertices on the proposed MPV model. (\textbf{a}) Synthetic image. (\textbf{b-e}) Vertices (yellow dots) and the corresponding constrained ADCs (blue lines) from the skeleton(red). (\textbf{f}) The predicted polygon mask by the PolarMask model. (\textbf{g-j}) Mask voting score maps.}
\label{fig_synInitial}	
\end{figure*}

In addition, the vertices $\fy_\ell$ predicted by the PolarMask model are often misaligned with the true object boundary in practical applications. Nevertheless, they can still be utilized as prescribed points to generate the constrained ADCs. By the method introduced in Section~\ref{subsec_DomainCuts}, we leverage each vertex $\fy_\ell$, the skeleton $\cI$ and the boundary $\partial\kS_0$ predicted by the PolarMask model to construct a constrained ADC $\cC_\ell$. This ADC is used to initialize the min-cut Randers geodesic model (see Section~\ref{subsec_Mincut}) to compute a circular geodesic path, denoted by $\cG_\ell$, from which we sample a set of new vertices $\sV_\ell=\{\fv_{\ell,i}\}_{1\leq i \leq I}$ with an approximately uniform spacing between successive vertices. Then the set $\sV$, used for the generation of mask proposals described in Section~\ref{subsec_DomainCuts}, can be constructed as 
\begin{equation}
\sV=\cup_{\ell}\sV_\ell \cup\{\fy_\ell\}_\ell.
\end{equation}
In experiments, the vertices $\fy_\ell$ for $1\leq \ell \leq L$ predicted by the PolarMask model are also included in $\sV$. We describe the procedure for the construction of the set $\sV$ in Algorithm~\ref{algo_Implementation}.

\begin{remark}
In the MPV model, constrained ADCs are constructed using a closed contour and a fixed vertex. For the ADCs corresponding to  vertices in the set $\{\fy_\ell\}_\ell$, predicted by the PolarMask model,  the closed contours is given by the mask boundary $\partial\kS_0$. In addition, for the set $\sV_\ell$, the  closed contour used is the circular minimal path $\cG_\ell$ which is also the contour serves as the initialization for the min-cut Randers model. Such an implementation ensures that both learning-based predictions and geometry-driven contours are consistently integrated into the ADC construction, thereby combining the advantages of data-driven and model-based approaches.
\end{remark}

\section{Experiments}
\label{sec_Experi}

In this section, we conduct qualitative and quantitative experiments to evaluate the performance of the proposed model against three representative minimal path-based approaches: the combination of the piecewise geodesic path model (CombPaths)~\cite{mille2015combination}, the asymmetric quadratic metric-based geodesic voting model (AsyVoting)~\cite{zhou2025generalized}, and the region-based Randers geodesic model (RegionGeo)~\cite{chen2024randers}. 
The evaluation is performed on both synthetic and real images through qualitative and quantitative comparisons. For quantitative assessment, the Dice index $\mathbb{D}$ is used to evaluate segmentation performance, defined as:
\begin{equation}
\label{eq:accuracyscore}
\mathbb{D}(S,G)=\frac{2|S\cap G|}{|S|+|G|},
\end{equation}
where $S$ denotes the segmented region, $G$ represents the ground truth, and $|\cdot|$ indicates the area of the region. The value of Dice index $\mathbb{D}$ ranges from 0 to 1, with higher values indicating better performance on the target segmentation. 

\subsection{Initialization for the Evaluated Models}
\label{sec_InitialTest}

In the proposed MPV model, the initialization data (i.e., the set of vertices $\sV$ and the skeleton structure $\cI$) is derived using the PolarMask model~\cite{xie2021polarmask++}. During training, the COCO benchmark~\cite{lin2014microsoft} is employed to train PolarMask for synthetic and natural image detection, while the Automated Cardiac Diagnosis Challenge (ACDC) dataset~\cite{olivier2018deep} is used for training on medical image segmentation tasks. In this work, we set the PolarMask model to predict $36$ vertices as the nodes of the polygon to model the segmentation mask. A subset of these vertices is selected to construct the required vertices $\sV$, with the number specified in each experimental setting.
For computing the constrained ADCs, the potential functions $\psi$ and $\psi_k$ are obtained using parameters $\tau = 3$, $w = 0.1$, and $\epsilon = 1$, see Eqs.~\eqref{eq_CutPotential} and~\eqref{eq_CutPotential2}. In addition, we set the width of the tubular neighborhood $\cT$ used in Eq.~\eqref{eq_CutPotential2} as $12$ grid points. The mask voting score map is generated by setting $\alpha_1 = 2$ and $\alpha_2 = 1$, as formulated in Eq.~\eqref{eq_Regiondetct}.
The ComPaths model extracts a closed contour by connecting user-provided points via piecewise geodesic paths, with a fixed maximum of $4$ admissible paths between successive vertices across all experiments.
In contrast, the AsyVoting model initializes with a single landmark point inside the target region. First, an adaptive cut is computed from this landmark. Next, a set of 500 endpoints is generated using a farthest-point sampling scheme from the adaptive cut. Minimal paths are then tracked using an asymmetric quadratic metric, yielding a geodesic voting map to delineate the boundary. 
Unlike the region-based Randers geodesic model~\cite{chen2024randers}, which requires boundary landmarks to construct an initial closed contour, our approach leverages only an interior landmark point and an adaptive cut to compute a circular isotropic geodesic path for initializing the Randers geodesic model. This simplifies user input while maintaining precision in boundary extraction.

\subsection{Results on Synthetic Images}
\label{sec_QualitComp}

\begin{figure}[h]
\centering
\includegraphics[width=0.95\linewidth]{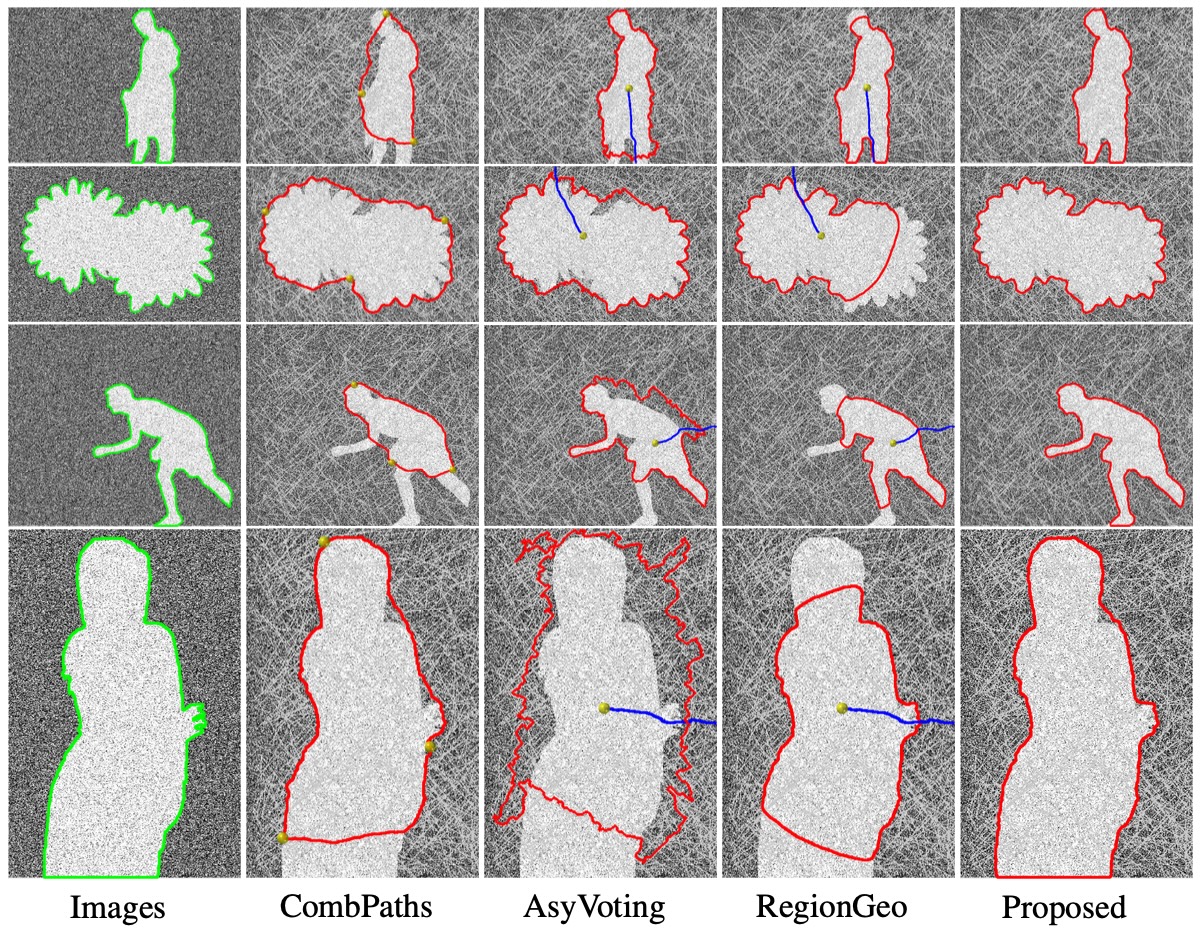}
\caption{Segmentation results on synthetic images corrupted by $n=700$ straight-line segments. The red lines represent the segmentation contours, the landmark points are denoted by yellow dots and the blue dash lines stand for the adaptive cuts. \textbf{Column 1}: The original synthetic images and the ground truth boundaries denoted by green lines. \textbf{Columns 2-5}: The closed contours extracted from the CombPaths model, the AsyVoting model, the RegionGeo model and the proposed MPV model, respectively.}
\label{fig:synNoiseFig}
\end{figure}

In this section, we conduct both qualitative and quantitative experiments on synthetic images to evaluate the performance of different segmentation models. The synthetic images are generated based on ground truth masks from the dataset in~\cite{rother2004grabcut}, where the pixel intensities of the target structures and the background are set to $1$ and $0.5$, respectively.
To simulate noise conditions, additive Gaussian noise with a normalized standard deviation of $\sigma_n=0.1$ is applied to each image, resulting in a set of $50$ synthetic images. To further challenge the segmentation algorithms, each image is corrupted by a number of randomly placed straight-line segments. The number of such segments varies across seven levels (from $n=100$ to $700$), with each segment having a length of $80$ pixels and an intensity value of $0.8$. These straight-line segments are able to produce strong edge saliency features, thereby increasing the difficulty of boundary extraction. Examples of synthetic images at the highest degradation level ($n=700$) are shown in Figs.~\ref{fig_synInitial} and~\ref{fig:synNoiseFig}.

\begin{table}[h]
  \centering
  \fontsize{7}{10}\selectfont
  \begin{threeparttable}
  \setlength{\tabcolsep}{6pt}
\renewcommand{\arraystretch}{1.1}
 \caption{Segmentation accuracy of the proposed model with varying keypoints.}
      \label{synInitial}
    \begin{tabular}{ccccccccc}
        \toprule
    \multirow{2}{*}{Keypoints}&
    \multicolumn{2}{c}{Synthetic images ($n=0$)}&\multicolumn{2}{c}{Synthetic images ($n=700$)}\cr
    \cmidrule(lr){2-3} \cmidrule(lr){4-5}      &Mean&Std&Mean&Std\cr
    \midrule
    8  &0.9609 &0.1406 &0.9510 &0.1407 \cr
    16 &0.9789 &0.0257 &0.9707 &0.0315  \cr
    24 &0.9860 &0.0072 &0.9734 &0.0247  \cr
    32 &0.9805 &0.0215 &0.9710 &0.0309  \cr
    \bottomrule
    \end{tabular}
    \end{threeparttable}
\end{table}

\renewcommand{\arraystretch}{1.5}
\begin{table}[t]
  \centering
  \fontsize{7}{10}\selectfont
  \begin{threeparttable}
  \setlength{\tabcolsep}{4.5pt}
\renewcommand{\arraystretch}{1.1}
\caption{Quantitative results under different numbers of straight-line segments 
$n$ used to generate the corruptions.}
     \label{tab:synNoise}
    \begin{tabular}{ccccccccc}
        \toprule
    \multirow{2}{*}{$n$}&
    \multicolumn{2}{c}{CombPaths}&\multicolumn{2}{c}{AsyVoting}&\multicolumn{2}{c}{RegionGeo}&\multicolumn{2}{c}{Proposed}\cr
    \cmidrule(lr){2-3} \cmidrule(lr){4-5} \cmidrule(lr){6-7} \cmidrule(lr){8-9}     &Mean&Std&Mean&Std&Mean&Std&Mean&Std\cr
    \midrule
    0   &0.8883 &0.1437 &0.9387 &0.1249 &0.9602 &0.0755  &{\bf0.9860}&{\bf0.0072}\cr
    100 &0.8895 &0.1500 &0.9355 &0.1301 &0.9529 &0.0626  &{\bf0.9833}&{\bf0.0130}\cr
    200 &0.8668 &0.1678 &0.9167 &0.1514 &0.9481 &0.0715  &{\bf0.9819}&{\bf0.0146}\cr
    300 &0.8854 &0.1456 &0.9175 &0.1602 &0.9340 &0.1470  &{\bf0.9803}&{\bf0.0172}\cr
    400 &0.8703 &0.1571 &0.8895 &0.1589 &0.9286 &0.0992  &{\bf0.9783}&{\bf0.0186}\cr
    500 &0.8622 &0.1678 &0.8762 &0.1794 &0.9175 &0.1487  &{\bf0.9760}&{\bf0.0234}\cr
    600 &0.8330 &0.1880 &0.8737 &0.1761 &0.8954 &0.1990  &{\bf0.9748}&{\bf0.0233}\cr
    700 &0.8494 &0.1782 &0.8367 &0.1961 &0.8849 &0.1997  &{\bf0.9734}&{\bf0.0247}\cr
    \bottomrule
    \end{tabular}
    \end{threeparttable}
\end{table}

\begin{figure}[t]
\centering
\includegraphics[width=0.49\textwidth]{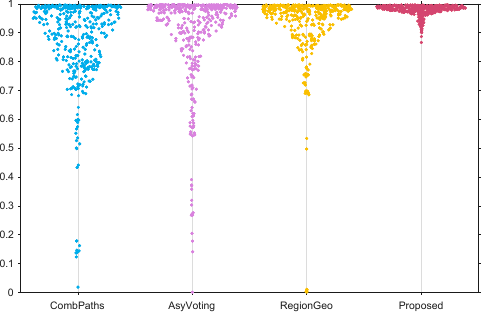}
\caption{Scatter plot of the Dice index over all synthetic images for different method.}
\label{fig:synNoiseScatter}	
\end{figure}

We evaluate the robustness of the proposed model by conducting experiments on synthetic images with degradation levels $n=0$ and $700$, focusing on the impact of keypoint quantity (ranging from $8$ to $32$ in four increments) along the target boundary for the voting scheme.
Starting from a pre-segmentation mask (Fig~\ref{fig_synInitial}f), four levels of vertices ($1$ to $4$) are selected. For each vertex, an orientation-constrained domain cut (blue lines) is computed from the skeleton structure (red lines), followed by a circular initial contour.  Eight keypoints are uniformly sampled near the target boundary as illustrated by yellow points in row $1$ of Fig.~\ref{fig_synInitial}. Subsequently, the proposed region proposal voting model is applied to generate the geodesic voting score maps via region-based voting, as shown in row $2$ of Fig.~\ref{fig_synInitial}. 
In addition, the effectiveness of the orientation-constraint domain cut can be demonstrated as well. The proposed orientation-constrained cuts are uniformly distributed across the image domain, enabling a comprehensive and robust representation of the target structure by capturing its geometric variability through multiple directional perspectives.
Quantitative results are reported using the Dice similarity coefficient to evaluate segmentation accuracy. The mean and standard deviation of the Dice scores under different numbers of keypoints are summarized in Table~\ref{synInitial}. The results show that the segmentation performance remains stable across different keypoint counts. Optimal performance is achieved with $24$ keypoints (highest mean Dice, low std), which is adopted for subsequent experiments.

The experiments on synthetic images degraded by straight-line segments of different levels is conducted to further demonstrate the robustness of the proposed method to noise. The average and standard deviation values of the Dice index under different levels of noises for all the considered models are illustrated in Table~\ref{tab:synNoise}.  It shows that the Dice index values derived from the proposed model are slightly affected by different kinds of noise with different levels, and have the best performance among all the compared models. We can see that the proposed model achieves the best robustness performance against noise, benefitting from region-based information into the geodesic voting model and incorporating global structural priors provided by the skeleton structure into the voting process. In addition, compared with the AsyVoting model with $500$ vertices for computing the geodesic paths, the proposed model obviously reduced the requirement to the number of vertices on the image.

To further assess the robustness of the proposed MPV model in comparison with the CombPaths, AsyVoting, and RegionGeo models, we conducted the experiments on synthetic images corrupted by varying densities of straight-line segments.
 The average and standard deviation of the Dice index for all evaluated methods under different noise levels are summarized in Table~\ref{tab:synNoise},  while the corresponding scatter plot in  Fig.~\ref{fig:synNoiseScatter} shows the Dice scores across all noised samples. As illustrated, the proposed MPV model consistently achieves the highest Dice scores with  lowest variance across all noise levels, indicating strong robustness to structured noise.
 This superior performance is attributed to the integration of regional  information into the geodesic voting framework, as well as the incorporation of global structural priors via the extracted skeleton. Furthermore, compared to the AsyVoting model, which requires 500 keypoints for geodesic path computation, the proposed model substantially reduces the reliance on vertex quantity while maintaining high segmentation accuracy. 

\begin{figure}[t]
\centering	
\includegraphics[width=0.95\linewidth]{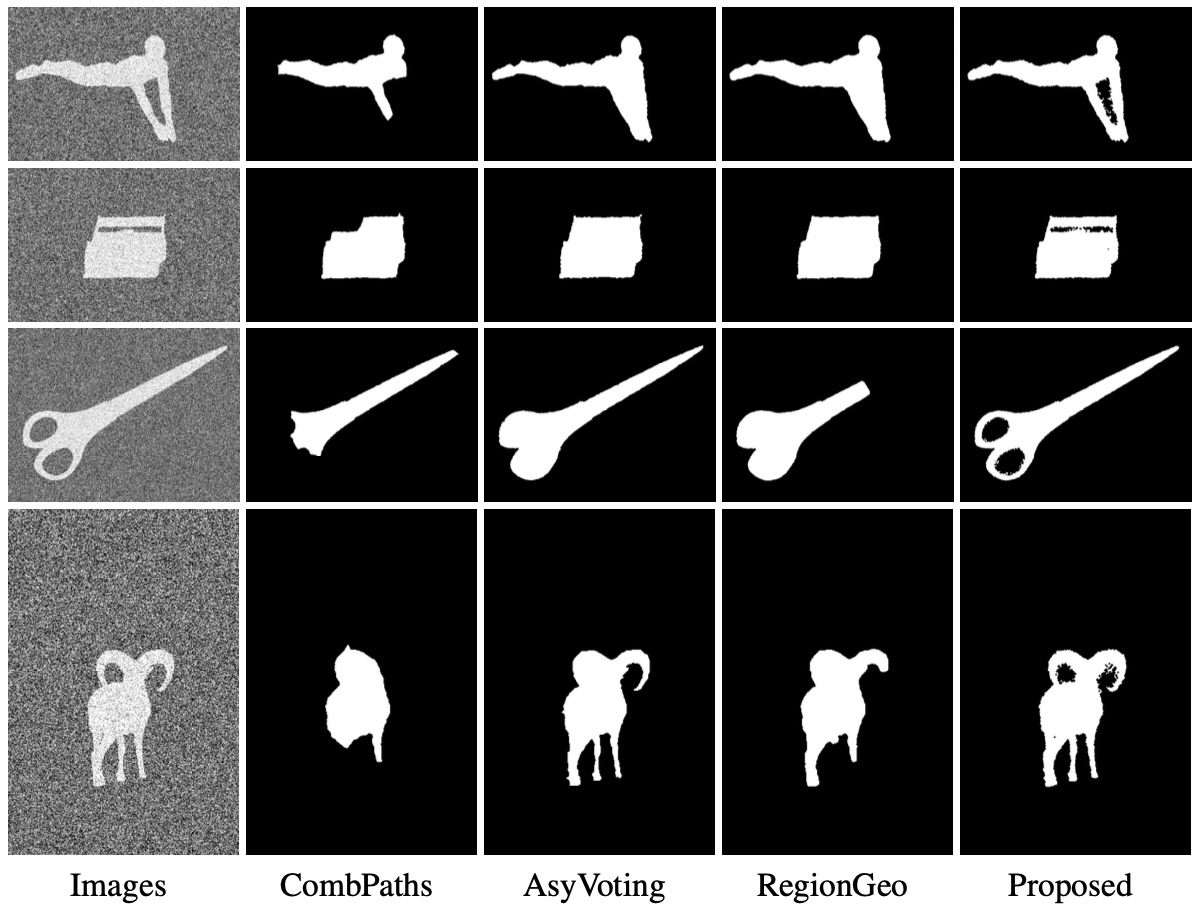}
\caption{Segmentation results on synthetic images interrupted by additive Gaussian noise. \textbf{Column 1}: Original synthetic images.  \textbf{Columns 2-5}: The segmentation region extracted from the CombPaths model, the AsyVoting model, the RegionGeo model and the Proposed model, respectively.}
\label{fig:synRegion}
\end{figure}

In Fig.~\ref{fig:synNoiseFig}, we qualitatively compare the segmentation performance of the evaluated models on synthetic images corrupted by $n=700$ straight-line segments. The segmentation results produced by the CombPaths model, the AsyVoting model, the RegionGeo model and the proposed MPV model are shown in Columns $2$ through $5$ of Fig.~\ref{fig:synNoiseFig}, respectively. In these visualizations, the adaptive cuts are illustrated by blue dashed lines, the ground truth boundaries are marked in green, the extracted contours are shown in red, and the landmark points are indicated by yellow dots. 
As observed, the CombPaths, AsyVoting, and RegionGeo models tend to miss fine structural details of the target object. The CombPaths model shows sensitivity to the number and placement of landmark points, requiring more comprehensive annotations along the boundary to achieve satisfactory results. The AsyVoting model is susceptible to strong edge saliency introduced by the synthetic straight-line noise, often resulting in noisy and irregular contours. The RegionGeo model, in contrast, may converge to incorrect regions due to local ambiguities.
In comparison, the proposed method effectively extracts the complete region of interest, generating a closed contour guided by the geodesic voting score map. This demonstrates superior robustness and structural consistency under challenging noise conditions.

\renewcommand{\arraystretch}{1.5}
\begin{table}[t]
  \centering
  \fontsize{7}{10}\selectfont
  \begin{threeparttable}
  \setlength{\tabcolsep}{4pt}
\renewcommand{\arraystretch}{1.1}
      \caption{Comparison of CombPaths, AsyVoting, RegionGeo, and Proposed models on the ACDC dataset~\cite{olivier2018deep}.}
       \label{tab:ACDC}
    \begin{tabular}{ccccccccc}
        \toprule
    \multirow{2}{*}{Dataset}&
    \multicolumn{2}{c}{CombPaths}&\multicolumn{2}{c}{AsyVoting}&\multicolumn{2}{c}{RegionGeo}&\multicolumn{2}{c}{Proposed Model}\cr
    \cmidrule(lr){2-3} \cmidrule(lr){4-5} \cmidrule(lr){6-7} \cmidrule(lr){8-9}     &Mean&Std&Mean&Std&Mean&Std&Mean&Std\cr
    \midrule
    ACDC   &0.8382 &0.1090 &0.4996 &0.3319 &0.6176 &0.3900  &{\bf0.9721}&{\bf0.0144}\cr
    \bottomrule
    \end{tabular}
    \end{threeparttable}
\end{table}

\begin{figure}[ht]
\centering
\includegraphics[width=0.95\linewidth]{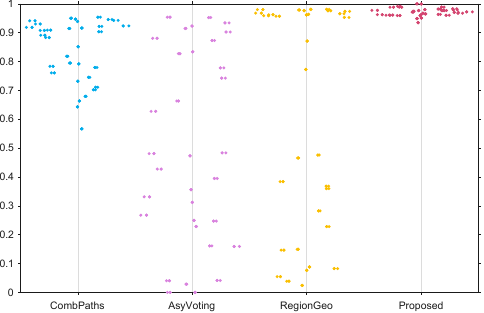}
\caption{Scatter plot of the Dice index over ACDC images for different method.}
\label{fig:synACDCScatter}	
\end{figure}

\begin{figure*}[t]
\centering
\includegraphics[width=0.75\linewidth]{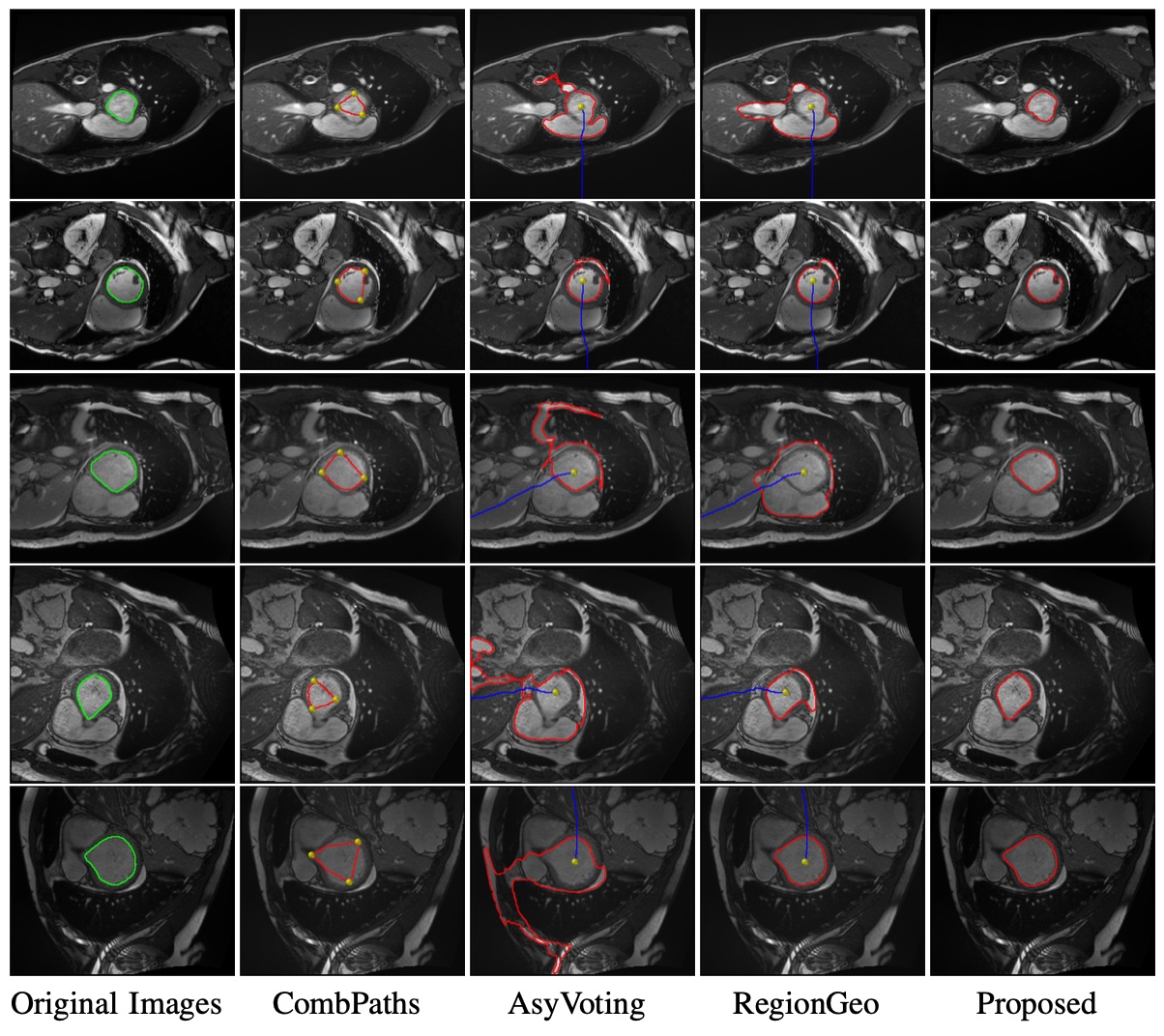}
\caption{Segmentation results on ACDC images. The red lines represent the segmentation contours and the landmark points are denoted by yellow dots. The blue dash lines stand for the cuts.  \textbf{Column 1}: Ground truth boundaries delineated on original images. \textbf{Columns 2-5}: The closed contours extracted from the CombPaths model, the AsyVoting model, the RegionGeo model and the Proposed model, respectively.}
\label{fig:ACDC}
\end{figure*}

In Fig.~\ref{fig:synRegion}, we illustrate the advantage of the proposed MPV  model over existing methods in segmenting target objects that contain internal background regions. The original synthetic images are shown in Columns 1 of Fig.\ref{fig:synRegion}. The segmentation results produced by the CombPaths, AsyVoting, RegionGeo, and the proposed models are presented in Columns 2 through 5 of Fig.\ref{fig:synRegion}.
As observed, the evaluated minimal path-based approaches, including CombPaths, AsyVoting, and RegionGeo, tend to misclassify background regions embedded within the target object as foreground, leading to structural segmentation errors. Both CombPaths and AsyVoting rely primarily on edge-based features and construct a series of minimal paths to delineate object boundaries. These methods, however, lack the capability to distinguish between interior background and true foreground regions. Although RegionGeo incorporates region-based data similarity/dissimilarity to generate a closed geodesic contour, it cannot simultaneously segment internal background areas due to its single-contour formulation.
In contrast, the proposed method achieves more accurate segmentation by leveraging a weighted region-based  voting strategy to construct a robust geodesic voting score map. The shape gradient information computed within each region proposal quantifies the likelihood of each point belonging to the interior or exterior region. Consequently, higher values in the voting score map correspond to a greater probability of being part of the true target region, enabling the model to more effectively separate foreground from internal background structures.

\begin{figure*}[t]
\centering
\includegraphics[width=0.75\linewidth]{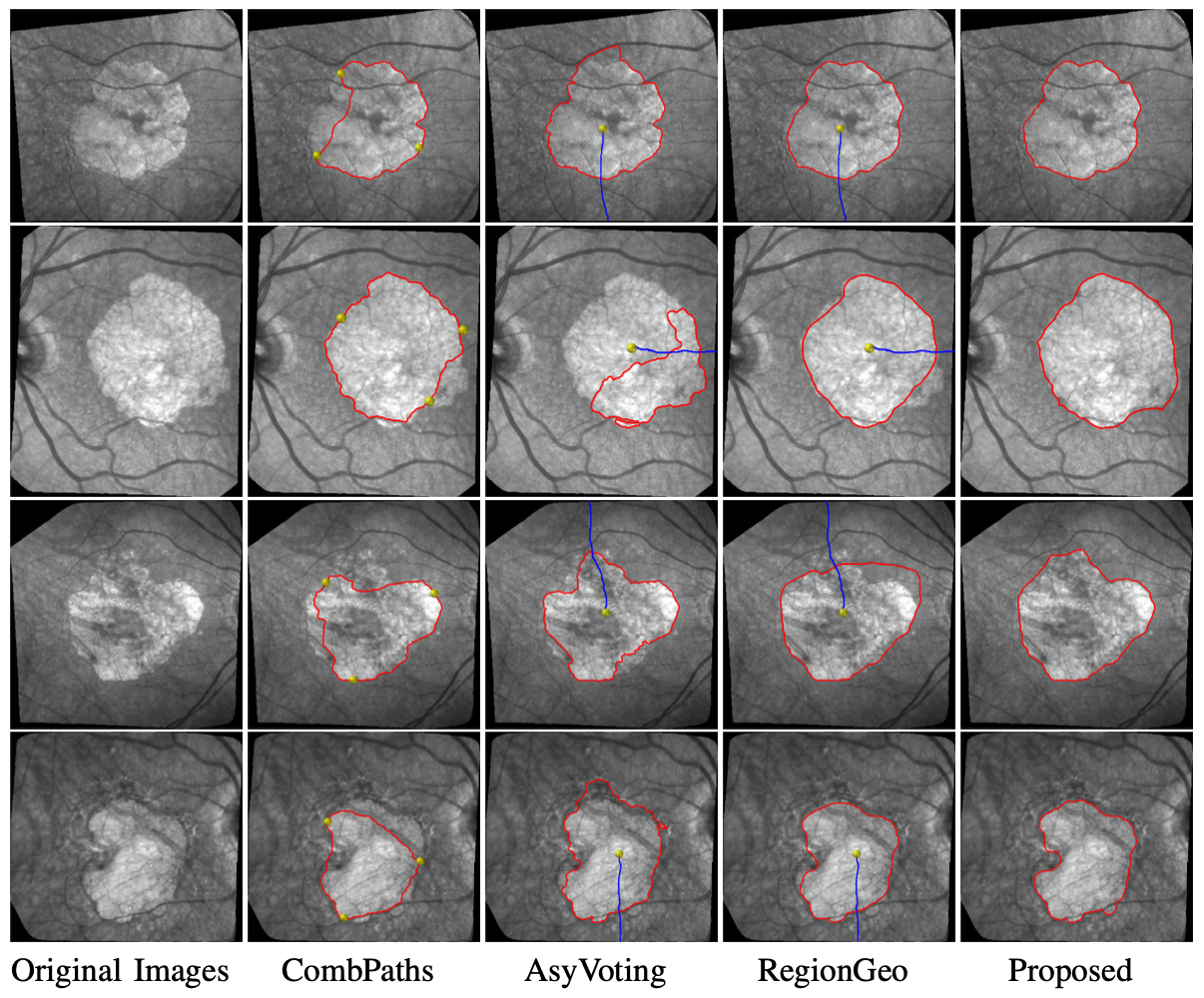}
\caption{Segmentation results on retina images. The red lines represent the segmentation contours and the landmark points are denoted by yellow dots. The blue dash lines stand for the cuts.  \textbf{Column 1}: Original images. \textbf{Columns 2-5}: The closed contours extracted from the CombPaths model, the AsyVoting model, the RegionGeo model and the Proposed model, respectively.}
\label{fig:retina}
\end{figure*}

\subsection{Results on Real Images}
\label{sec::QuantiComp}

In this section, we evaluate the segmentation performance of the proposed method and baseline models on real medical images. Specifically, qualitative and quantitative experiments are conducted using $30$ MRI slices from the Automated Cardiac Diagnosis Challenge (ACDC) dataset~\cite{olivier2018deep}, with the objective of segmenting the left ventricle.
In the proposed method, the number of keypoints used to construct the geodesic voting score map is fixed at 8. The mean and standard deviation of the Dice index for all compared methods are reported in Table~\ref{tab:ACDC}, while a scatter plot of individual DSC values across the $30$ test images is illustrated in Fig.~\ref{fig:synACDCScatter}.
The results demonstrate that the proposed method consistently achieves the highest segmentation accuracy, exhibiting both the highest average Dice score and the lowest standard deviation among all competing models. This indicates the superior robustness and precision of our approach in delineating cardiac structures from real MRI data.

\begin{figure}[t]
\centering
\includegraphics[width=0.9\linewidth]{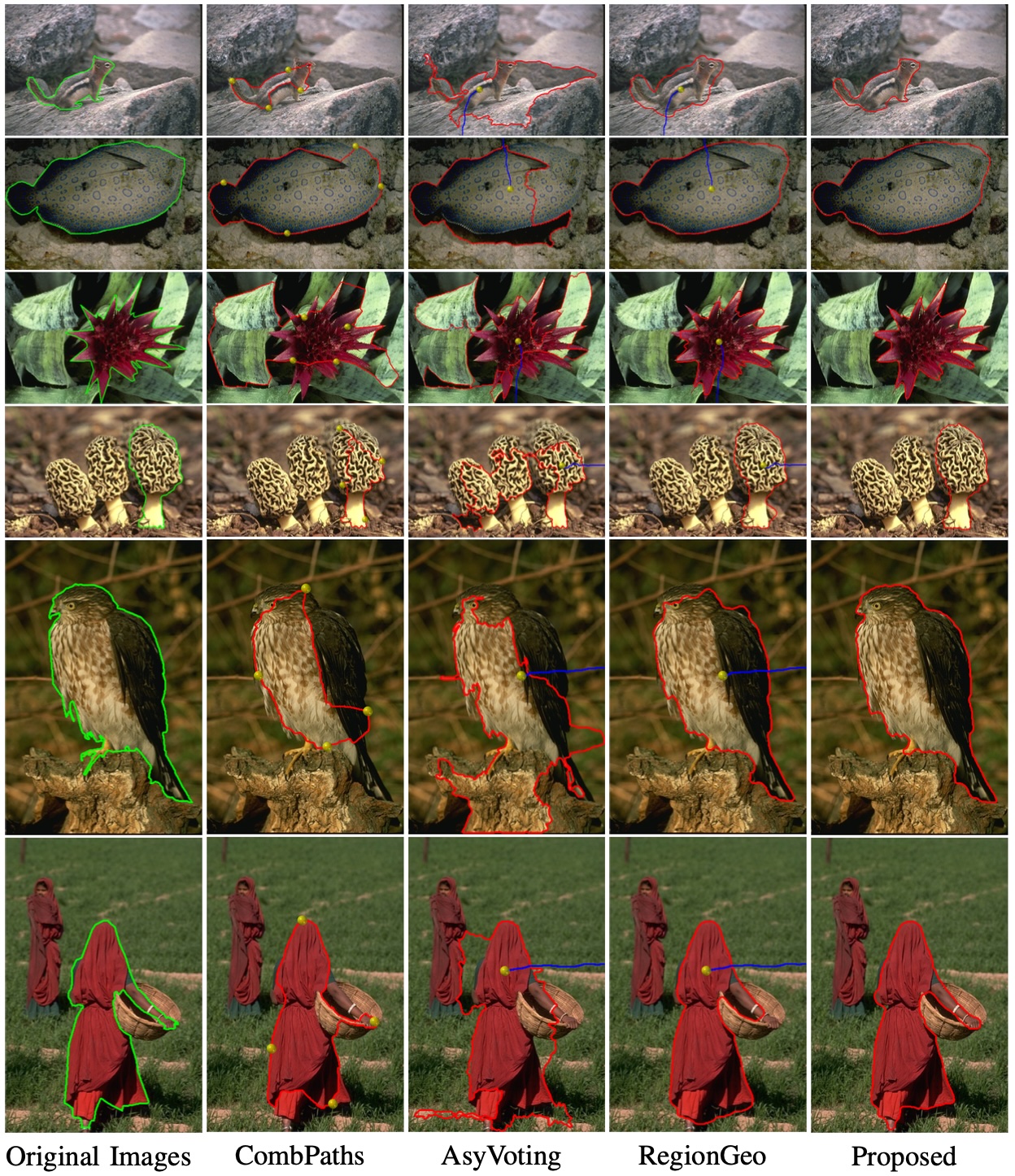}
\caption{Segmentation results on nature images. The red lines represent the segmentation contours and the landmark points are denoted by yellow dots. The blue dash lines stand for the cuts.  \textbf{Column 1}: Original images with ground truth boundary. \textbf{Columns 2-5}: The closed contours extracted from the CombPaths model, the AsyVoting model, the RegionGeo model and the Proposed model, respectively.}
\label{fig:nature}
\end{figure}

In addition to the quantitative results, qualitative comparisons of the evaluated models on representative ACDC MRI images are presented in Fig.~\ref{fig:ACDC}. These images exhibit challenges such as shape variability, small target structures, and complex surrounding backgrounds.
In Fig.~\ref{fig:ACDC}, ground truth boundaries are marked in green, predicted contours in red, adaptive domain cuts in blue, and landmark points in yellow. Column $1$ depicts the original images and the corresponding ground truth boundaries (green lines). Column $2$ displays the segmentation results produced by the CombPaths model, where inaccuracies can be observed near regions with blurred object boundaries. Column $3$ shows the results from the AsyVoting model, which is notably affected by background clutter and local edge inconsistencies. Column $4$ presents the results from the RegionGeo model. However, in cases of complex backgrounds, it becomes difficult to accurately select the intersection points between the adaptive cut and object boundary, leading to suboptimal geodesic path tracking.
In contrast, the proposed model (Column $5$) successfully delineates the target structures. This improvement is enabled by the integration of pre-computed shape and positional priors, which guide the geodesic voting mechanism and enhance robustness in challenging scenarios.

In Fig.~\ref{fig:retina}, we present a qualitative comparison of the aforementioned models on retinal images, where the objective is to segment lesion regions. The target structures typically exhibit inhomogeneous intensity distributions, are embedded in complex backgrounds, or possess blurred boundaries. The segmentation results are depicted as red contours, while the adaptive cuts and landmark points are represented by blue dashed lines and yellow dots, respectively. 
Column $1$ displays the original images. Columns $2$ and $3$ show the results produced by the CombPaths and AsyVoting models, respectively. As both rely predominantly on edge features, they struggle to achieve satisfactory segmentation performance in the presence of inhomogeneous regions. Columns $4$ and $5$ illustrate the outputs of the RegionGeo and proposed models. Both methods leverage region-based cues and are capable of delineating most parts of the target structures. However, the proposed model, benefiting from the region proposal voting strategy, achieves more refined and accurate segmentation results, particularly in challenging inhomogeneous areas.

In Fig.\ref{fig:nature}, we present a qualitative comparison on natural images sampled 
from the dataset~\cite{martin2001database}. The segmentation results of the CombPaths, AsyVoting, RegionGeo, and proposed models are shown in Columns 2 through 5, respectively. The adaptive cuts initiated from internal points are illustrated by blue dashed lines, extracted contours are marked in red, and landmark points are indicated by yellow dots.
As observed, the CombPaths model tends to follow boundaries with strong edge features, often resulting in incomplete segmentation of the target objects. Unlike CombPaths, which uses four landmark points to constrain the boundary, the AsyVoting model attempts to trace a closed contour from an adaptive cut but may incorrectly include background regions due to insufficient boundary localization. While the RegionGeo model can roughly localize the target, it often fails to accurately delineate the boundary, particularly in cluttered scenes. In contrast, the proposed model demonstrates superior performance by effectively extracting the region of interest as a complete and accurate closed contour, benefiting from the integration of region proposal voting and structural priors.

\section{Conclusion and Discussion}
\label{Sec_Conclusion}
In this paper, we present a novel segmentation framework, termed the \emph{mask proposal voting model}, which integrates constrained ADCs with a mask proposal voting scheme with different importance weighting assignment. A central component of this framework is the computation of constrained ADCs, guided by the skeleton of the target region, which provides efficient initialization for the min-cut geodesic model. Mask proposals are subsequently generated using the ADC-driven min-cut Randers geodesic model based on a set of landmark points, enabling robust extraction of complex and inhomogeneous target regions. Extensive experiments on synthetic, medical, and natural images demonstrate the superior accuracy and robustness of the proposed method, particularly in challenging scenarios characterized by weak boundaries and complex image content.

Future work will focus on (i) incorporating curvature regularization and curvature-inspired shape priors~\cite{chen2023computing,chen2017global,mirebeau2018fast,mashtakov2023time,van2024geodesic} into the MPV model, and (ii) extending the current two-dimensional framework to three-dimensional image segmentation~\cite{10462911,10930660}. In particular, we aim to adapt the ADC and geodesic voting mechanisms to volumetric data, thereby facilitating more accurate delineation of anatomical structures in 3D medical imaging applications such as CT and MRI volumes.



\bibliographystyle{IEEEtran}
\bibliography{minimalPaths}

\ifCLASSOPTIONcaptionsoff
  \newpage
\fi

\end{document}